# Document stream clustering: experimenting an incremental algorithm and AR-based tools for highlighting dynamic trends.

ALAIN LELU[1]
MARTINE CADOT[2]
PASCAL CUXAC[3]

1 LASELDI / Université de Franche-Comté
30 rue Mégevand – 25030 Besançon cedex
alain.lelu@univ-fcomte.fr
2 UHP/LORIA Campus Scientifique BP 239 - 54506 Vandoeuvre-lès-Nancy Cedex
martine.cadot@loria.fr
3 INIST/CNRS 2 Allée du Parc de Brabois - CS 10310 - 54514 Vandoeuvre-lès-Nancy Cedex
pascal.cuxac@inist.fr

*Abstract*

We address here two major challenges presented by dynamic data mining: 1) the stability challenge: we have implemented a rigorous incremental density-based clustering algorithm, independent from any initial conditions and ordering of the data-vectors stream, 2) the cognitive challenge: we have implemented a stringent selection process of association rules between clusters at time t-1 and time t for directly generating the main conclusions about the dynamics of a data-stream. We illustrate these points with an application to a two years and 2600 documents scientific information database.

## *1. Introduction*

Data-stream clustering is an ever-expanding subdomain of knowledge extraction. Most of the past and present research effort aims at efficient scaling up for the huge data repositories, see Gaber et al [1]. In our view, a fully grounded scientific approach needs a prior focus upon the quality and stability of the clustering - in a first phase, on relatively small scale and humanly apprehensible datasets, then scaling up. Our approach insists on reproducibility and qualitative improvement, mainly for "weak signals" detection and precise tracking of topical evolutions in the framework of information watch: our GERMEN algorithm exhaustively picks up the whole set of density peaks of the data at time t, by identifying the local perturbations induced by the current document vector, such as changing cluster borders, or new/vanishing clusters.

However, this is only one side of the medal: on the user side of the problem, it is of the utmost importance to provide him/her with tools for synthesizing the dynamic information in a humanly perceptible form, so that he/she may quickly apprehend the main tendencies in the data-flow. This is a big challenge indeed, since the *simpler* problem of representing a static multidimensional phenomenon has generated huge amounts of literature in the past forty years, in fields so diverse as statistic data analysis, text mining, data visualization, graph mining, knowledge extraction and representation…

The present paper is meant to present a few modest steps in this direction.

One evident way for tackling the problem may be to introduce moving graphical representations – shifting from still pictures to movie films, so to speak.

We explore here another perspective: human language being itself a dynamic process, it has a powerful expression capability for describing dynamic phenomena. Sentences such as "A new trend emerged from the 2005 data", "The formerly cohesive group A started to evenly decay in the early 2000'", or "in 2004 the group B split into two distinct subgroups" are the type of information human brains can directly assimilate and process. This kind of presentation is not far from a rule-based one. In this paper we try to explore fuzzy Association Rules (AR) as a way for directly generating such tendency reports.

State-of-the-art ARs are often sketched as "generating more data than the data they are supposed to mine"... This is not far from true, and this is why 1) we have intensively investigated statistical methods for skimming and validating the most powerful and non-redundant fuzzy rules sets, 2) our rules are applied to the *restricted* number of entities brought to light in the dynamic clustering phase, not to the raw data.

The first section will develop and argue for the principles of our incremental clustering algorithm, more precisely described in Lelu [2] and Lelu, Cuxac [3]. The next section will present more extensively our work on extracting, minimizing and validating sets of our fuzzy association rules described in Cadot et al. [4]. Both sections will be illustrated by our experimental work on a subset of Pascal scientific information database (CNRS/INIST), relative to the geotechnics field along the two years 2003-2004 (~2600 records).

## 2. *A wish list for a data stream clustering algorithm, and an experimental implementation and application*

### 2.1. *State of the art and discussion*

A *precise* and fine-grained track for evolutions in a data-flow is needed in many application domains, particularly information watch ("weak signals" hunting…). From our experience in text-mining clustering and visual exploration tools, we feel it necessary:

1) to build a dynamic data-stream analysis upon a *stable* and reproducible static clustering process, i.e.:

- independent from the ordering of the data vectors (requirement #1),

- independent from any initial condition, which discards all the state-of-the-art methods converging to a local optimum of a global quality criterion (k-means and variants: Binztock, Gallinari [5], Chen et al. [6], as well as EM-based clustering methods, as presented in Buntine [7]) (requirement #2),

- needing a minimum number of parameters, if not any, for restraining the combinatorics of possible choices, and maximizing verifiability and reproducibility (requirement #3).

2) to require a fully incremental implementation, in order to catch fine-grained temporal evolutions, such as the emergence of new trends.

Out of the few basic clustering principles (hierarchical, centroïds, …) only one, in our view, seems to satisfy the whole set of these requirements: density-based clustering, such as Trémolières [8][9], Ester et al. [10], Moody [11], Guénoche [12], Hader et Hamprecht [13], Batagelj [14]; we have added to the elegant one-parameter percolation method of Trémolières:

- an adaptive definition of density, as used by Ertöz et al. [15], based on the K-Nearest-Neighbours definition of neighbourhood.

- the incremental and distributed features, inspired by the adaptive networks protocols in "Ad-hoc" radiocommunications field, initiated by Mitton, Fleury [16].

## 2.2. Our incremental density-based algorithm

This algorithm is fully described and commented in Lelu [2] and Lelu, Cuxac [3]. The principles and pseudocode description are as follows:

After a non-conventional normalization of the document input vector, we update a K-nearest neighbors similarity graph between the prior input vectors, whose valued edges are the cosine values between neighbors; to each node is attributed a density value, derived from the well-known clustering coefficient, and one or several clusterhead IDs.

The new input vector induces local perturbations in the density values, or "density landscape" of the graph, and thus in the clusterhead structure: it may create a new cluster, and be itself clusterhead, if it is denser than its neighbors; or it may link up with existing one(s) and alter the influence areas of the clusterheads, by means of our multiple inheritance rule – a document-vector inherits the clusterhead IDs from all the "overhanging", denser nodes in its 1-neighborhood.

```
. Initialization: the first node in the sequence has no link, has a 0 density and is its own clusterhead.
LCC = Ø   // LCC is the list of clusterhead lists for each node //

. FOR each new node :

    // induced changes for densities: //
    .compute its 1- et 2-neighbourhood (income and outcome), and other induced neighbourhood modifications, from which
    follows the list LL of the nodes concerned by a link creation / suppr. / modif.
    .compute its density from its 1-neighbourhood.
    .FOR each LL node, and any node in its 1- neighbourhood:
        - compute the new density value
    End FOR

    // induced changes for clusterheads: //
    L = LL
    WHILE the list L of nodes akin to change their state is not empty :
        list LS = Ø
        FOR each L node, sorted by decreasing density :
            ~ apply the rule for clusterhead change according to clusterheads of the inbound denser neighbors (in LCC)
              and their densities.
            ~ if a change occurs:
                .update LCC for the current node
                .compute the possible overhanged nodes (overhanged by the current node);
                .increment the LS list of overhanged nodes.
        End FOR
        L=LS
    End WHILE
  end FOR
```

The choice of our unique parameter K has proved to be a compromise between excessive fragmentation of the clusters (K=1) and excessive chaining effects (K>=4).

## 2.3. Application to a 2-years evolution of geotechnical literature

PASCAL is a general science bibliographic database edited by CNRS / INIST. We have extracted 2598 records in the field of geotechnics, from 2003 (1541 papers) to 2004 (1057 papers), described by a vocabulary of 3731 keywords, once eliminated frequent generic or off-topic terms as well as rare ones.

Our GERMEN algorithm, with parameter K=3, created 179 kernels at the step 2003, 294 at the step 2004. Papers are distributed approximately as follows: 50% in the kernels, of size ranging from 2 to 35 papers, 8% in individual nodules, or "curds", 15% are outliers (individual kernels with no paper attached), 27% are linked to more than one kernel or curd (N-arity >1). These statistics illustrate how dispersed and sparse are our cluster structures.

Table 2 shows an example of a kernel of 7 papers and 12 "polysemic" papers attached to it, of N-arity 2 to 5.

Table 3 illustrates this kernel with the main relative participations of keywords to the intra-kernel links.

```
Kernel number (clusterhead) = 479: Stability monitoring

N-arity / internal # / density / external # / Title
!1   479   11.705035  017320 1   Geotechnical investigation of the New Baltimore Slide
!1   92     9.9605749 016933 1   Geotechnical trends in urban terrains evolution
!1   745    8.8692265 017586 1   Monitoring of Metro Line C construction in Prague
!1   161    8.7494076 017002 1   TAILSAFE: investigation and improvement of tailings facilities
!1   60     7.4491246 016901 1   Monitoring of the test on the dike at Bergambacht: design and practice
!1   725    5.8891947 017566 1   Monitoring and physical model simulation of a complex slope deformation in neovolcanics
!1   474    4.6110304 017315 1   Displacement measurement by image processing in a field test on dike failure
!
!2   142    8.1226061 016983 1   Methods of monitoring of historical buildings and slopes
!2   760    6.8718024 017601 1   Parametric analysis of slope stability analysis using AutoLISP
!2   226    5.9199853 017067 1   Uncoupled analysis of stabilizing piles in weathered slopes
!2   103    5.8558796 016944 1   Multi-method investigation of contaminated soils for a sustainable land use
!2   61     5.5634301 016902 1   An interdisciplinary approach for brownfields
!2   388    5.0722303 017229 1   Cumulation of seismic waves during formation of kimberlite pipes
!2   26     4.8760059 016867 1   Thermal strengthening of loess soils
!
!3   718    4.6264401 017559 1   Rock cleaning along Bavaria's road network
!3   478    4.1930455 017319 1   Instrumentation and monitoring of combined piled rafts (CPRF): state-of-the-art report
!3   653    3.8222772 017494 1   RN 88. Déviation de La Guide La Besse. Un chantier de terrassements complexes
!3   805    3.5239168 017646 1   Apparent phase velocities and fundamental-mode phase velocities of Rayleigh waves
!
!5   241    3.5936722 017082 1   La place de la géotechnique dans l'organisation des projets et des travaux
```

Table 1

```
‰ internal links / keyword # / keyword
!130  348    Surveillance        !
!90   601    Pile                !
!45   323    Loading test        !
!45   330    Seismic wave        !
!45   111    Surface wave        !
!34   39     Wave dispersion     !
!34   885    Models              !
!34   218    Mass movements      !
!34   355    S wave              !
!34   213    Lift                !
!34   528    Rupture             !
!34   66     Earthworks          !
!34   9      Phase velocity      !
!28   397    Loading             !
!28   1016   Slope stability     !
!22   14     Industrial wasteland !
!17   426    Numerical analysis  !
!11   146    Image analysis      !
!11   112    Freeway             !
!11   9      Safety coefficient  !
!11   173    Construction        !
!11   580    Displacement        !
```

Table 2

## 3. Fuzzy association rules for a limited and validated set of "hot conclusions" concerning the dynamics of a knowledge domain

### 3.1. Definition of association rules

Data mining techniques can be used to analyse large amounts of data (see Hand et al. [17]). Association Rules Extraction belongs to these techniques. It is a method for extracting knowledge relative to links between variables from data which can be represented by a boolean array of type O*V, where O is a set of objects, V a set of variables, and the boolean values indicate the presence or absence of a relation (such as belonging) between each variable and each object. The most widely cited association rule was extracted from US market basket data composed of transactions (lists of items bought in a single purchase by a customer): "x buys diapers → x buys beers" (Han et al. [18]).

In this market database, V is the item set, O the transaction set, and the values TRUE/FALSE indicate whether items are present or not in this transaction. Clearly, not all the customers who buy diapers also buy beers. But the proportion of customers that verify this surprising rule is sufficiently high to be interesting to supermarket managers. Presently, association rules extraction is not used to mine only marketing databases. It is an efficient data mining technique for extracting knowledge from most databases (Cadot et al [19]), for example corpuses of scientific texts as in the proposed application.

AR are commonly mined in two steps (Agrawal et al. [20]). First, new variables, called itemsets, are derived from the variables of the set V. For doing so, each object is affected a value of this new variable, say AB, by means of a logical AND on the concerned variables, A and B. The support of the itemset is the number of objects set to TRUE, in other words the number of "common" objects for the variables in the itemset. By fixing a threshold to the support value one can limit the number of extracted itemsets; those are called "frequent itemsets". The quest for powerful algorithms to do so is still a concern today. Most usual algorithms (Bastide [21]) are iterative ones (such as Agrawal « Apriori ») in which the knowledge of frequent itemsets of length k enables one to extract the ones of length k-1. The second step consists in splitting each itemset (k>1) into two item subsets A and B in order to test the association rule A→B. For limiting the number of generated rules, a threshold is applied to a quality criterion; the confidence (conf(A→B) = supp(AB) / supp(A)) is generally chosen. More than fifty quality criteria have been proposed yet (Guillet et al. [22]), each one embedding some specific semantic facet of the association rule. At last, the whole set of extracted AR is presented to the domain expert, jointly with some selected quality criteria.

This extraction method is currently enhanced along several directions: for time-efficiency, for taking into account more data-types, or more rule types. Specifically for non-boolean variables, e.g. with values scaling from 0 to 1, it is possible to define fuzzy itemsets by commuting the AND operator into a fusion operator adapted to real positive values (Cadot et al. [23]). This is the case for our clustering results on geotechnics data (Cuxac et al. [4]). We have then built fuzzy interclusters itemsets, which enabled us to induce fuzzy AR between our topics.

But many of the (too many!) rules derived from more than 2-items sets proved meaningless. This was one of our motivations for designing our MIDOVA index (Multidimensional Interaction Differential of Variables Association), so as to present to the expert the only rules built on itemsets showing an information content greater than their subsets (Cadot et al. [24]), thus eliminating useless redundancy. The numerical and dimensional properties of this index, formally similar to those of the support, allow for using an Apriori-like algorithm, for quickly extracting itemsets (whose MIDOVA absolute value reflects a strong association when the value is positive, a strong repulsion in the negative case). Most of the rules derived from itemsets with highly positive MIDOVA indices proved meaningful.

## 3.2. Use of association rules for comparing classes

Let C1 and C2 be two distinct classifications of two [documents × keywords] datasets sharing the same common keywords, A be a C1 class, B be a C2 class. Let us consider the matrix whose lines are the keywords (the O set), columns are the union V of the class labels of C1 and C2, and values are the typicalities of the keywords in each class, as defined above in 2.2.

The association between A and B may be considered as the itemset AB, i.e. a new column condensing the content of the two classes. The quality of the A→B rule depends on the « degree of inclusion » of A in B, as defined above by our notions of fuzzy itemsets and fuzzy rules.

In a first approximation, we will consider interesting the associations featured with a strictly positive MIDOVA coefficient. There are 236 of those in our dataset, issued from 182 C1 classes (upon 297 with one word at least) and 200 C2 classes(upon 497).

|   |   | B (C2) | | | Total |
|---|---|---|---|---|---|
|   | # premises | 1 | 2 | 3 |   |
| A (C1) | 1 | 114 | 12 | 1 | 127 |
|   | 2 | 20 | 6 |   | 26 |
|   | 3 | 4 | 1 |   | 5 |
|   | 4 |   |   | 1 | 1 |
|   | 5 | 1 |   |   | 1 |
| Total |   | 139 | 19 | 2 | 160 |

Table 3

In Table 3 we cross-count rules with 1,2,…5 premises issued from C1 classes along with rules with 1,…3 premises issued from C2 classes. This table shows four prominent types of associations:
- the "114" value means that there exist 114 AB "unique" itemsets, unique in that no itemsets can be found with the same class A from C1 and another class from C2 – which can be interpreted as a reciprocal coverage of A and B; of course we mean here a fuzzy coverage.
- "12" means that 12 C1 classes generate itemsets AB and AB', where B and B' are two C2 classes, thus generate 24 associations. We can interpret that as a split of A into two classes B and B'.
- "20" can be interpreted symmetrically as a fuzzy melting of two C1 classes A and A' into one C2 class B.
- "6" means that there are 6 C1 pairs (A, A') associated with 6 C2 pairs (B, B') into 4 itemsets AB, A'B, AB' and A'B', signalling cross-melting/splitting processes.

To summarize, 114 C1 (2003) classes stayed practically unchanged in 2004, 68 C1 classes combined by melting/splitting into 86 modified C2 (2003-2004) classes. The 115 C1 classes out of the table have no association with C2 classes, thus disappeared. The 297 C2 classes out of the table are newcomers.

This global comparison of the classifications C1 and C2 can be complemented with local fuzzy AR. We show here as an example two rules describing the split of a 2003 class ("Groundwater flow ") into two 2003-2004 classes ("Response of exploited underground water-table", "Groundwater flow "):

Rule(1) A03t1526→a34t2564 ; support : 6.65, MIDOVA : 3.29, confidence : 0.66
Rule(2) A03t1526→a34t0253 ; support : 5.74, MIDOVA : 1.47, confidence : 0.57

The high support and MIDOVA values show the strong similarity between the two pairs. The higher confidence in rule (1) is a sign of dissymmetry, the class A03t1526 being a bit more influenced in the direction of a34t2564.

In the same way, other noticeable examples may be cited:

- Example of merging classes: a03t0790 + a03t1017 → a34t0486

(Structures settlement + Reinforced works → Diagnosis of damaged works)

- Example of dying out: a03t1220 → []

(Shock wave in the grounds (except seismic wave) → [])

## *4. Conclusion*

Beyond the limits of the present options embedded in our algorithms, we have shown that the two major challenges posed by dynamic data mining could be addressed:

- the stability challenge: we have implemented a rigorous incremental density-based clustering algorithm, independent from any initial conditions and ordering of the data-vectors stream.

- the cognitive challenge: we have implemented a stringent selection and ranking process of association rules between clusters at time (t-1) and clusters at time (t) for directly generating the main conclusions about the dynamics of a data-stream.

These processes are independent and may be refined independently. Our further investigations will aim at scaling up and making the general granule size of our clustering process closer to human categorizing habits, as well as evaluating our rule's ranking criteria for enhancing the presentation of strong, indisputable rules first.

## *Acknowledgements*